\documentclass[lettersize,journal]{IEEEtran}
\usepackage{algorithmic}
\usepackage{algorithm}
\usepackage{array}
\usepackage[caption=false,font=normalsize,labelfont=sf,textfont=sf]{subfig}
\usepackage{textcomp}
\usepackage{stfloats}
\usepackage{url}
\usepackage{verbatim}
\usepackage{graphicx}
\usepackage{cite}
\usepackage{amsmath,amssymb,amsfonts}
\usepackage{enumitem}
\usepackage{pifont}
\usepackage{soul}
\usepackage[x11names,table,svgnames,xcdraw]{xcolor}
\usepackage{multirow}
\usepackage{booktabs}
\begin{document}

\title{Language-guided Scale-aware MedSegmentor for Lesion Segmentation in Medical Imaging}

\author{Shuyi Ouyang, Jinyang Zhang, Xiangye Lin, Xilai Wang, Qingqing Chen, Yen-Wei Chen, Lanfen Lin
        % <-this % stops a space
\thanks{Shuyi Ouyang, Jinyang Zhang, Xiangye Lin, Xilai Wang,  and Lanfen Lin are with the College of Computer Science and Technology, Zhejiang University,
Hangzhou 310058, China (e-mail: \{oysy, jinyang.zhang, xiangyelin, xilaiwang, llf\}@zju.edu.cn).}
\thanks{Qingqing Chen is with the
Department of Radiology, Sir Run Run Shaw Hospital, Hangzhou 310009, China (e-mail: qingqingchen@zju.edu.cn).}
\thanks{Yen-Wei Chen is with the College of Information Science and Engineering,
Ritsumeikan University, Kyoto 603-8577, Japan (e-mail: chen@is.ritsumei.ac.jp).}
}

% The paper headers
\markboth{Journal of \LaTeX\ Class Files,~Vol.~14, No.~8, August~2021}%
{S. OUYANG \MakeLowercase{\textit{et al.}}: Language-guided Scale-aware MedSegmentor for Medical Image Referring Segmentation}

\IEEEpubid{0000--0000/00\$00.00~\copyright~2021 IEEE}
% Remember, if you use this you must call \IEEEpubidadjcol in the second
% column for its text to clear the IEEEpubid mark.

\maketitle

\begin{abstract}
In clinical practice, segmenting specific lesions based on the needs of physicians can significantly enhance diagnostic accuracy and treatment efficiency. However, conventional lesion segmentation models lack the flexibility to distinguish lesions according to specific requirements. Given the practical advantages of using text as guidance, we propose a novel model, Language-guided Scale-aware MedSegmentor (LSMS), which segments target lesions in medical images based on given textual expressions. We define this as a new task termed Referring Lesion Segmentation (RLS). To address the lack of suitable benchmarks for RLS, we construct a vision-language medical dataset named Reference Hepatic Lesion Segmentation (RefHL-Seg).
LSMS incorporates two key designs: (i) Scale-Aware Vision-Language attention module, which performs visual feature extraction and vision-language alignment in parallel. By leveraging diverse convolutional kernels, this module acquires rich visual representations and interacts closely with linguistic features, thereby enhancing the model’s capacity for precise object localization. (ii) Full-Scale Decoder, which globally models multi-modal features across multiple scales and captures complementary information between them to accurately delineate lesion boundaries.
Additionally, we design a specialized loss function comprising both segmentation loss and vision-language contrastive loss to better optimize cross-modal learning.  We validate the performance of LSMS on RLS as well as on conventional lesion segmentation tasks across multiple datasets. Our LSMS consistently achieves superior performance with significantly lower computational cost. Code and datasets will be released.
\end{abstract}


\section{Introduction}

\begin{figure}[t!]
  \centering
  \includegraphics[width=\linewidth]{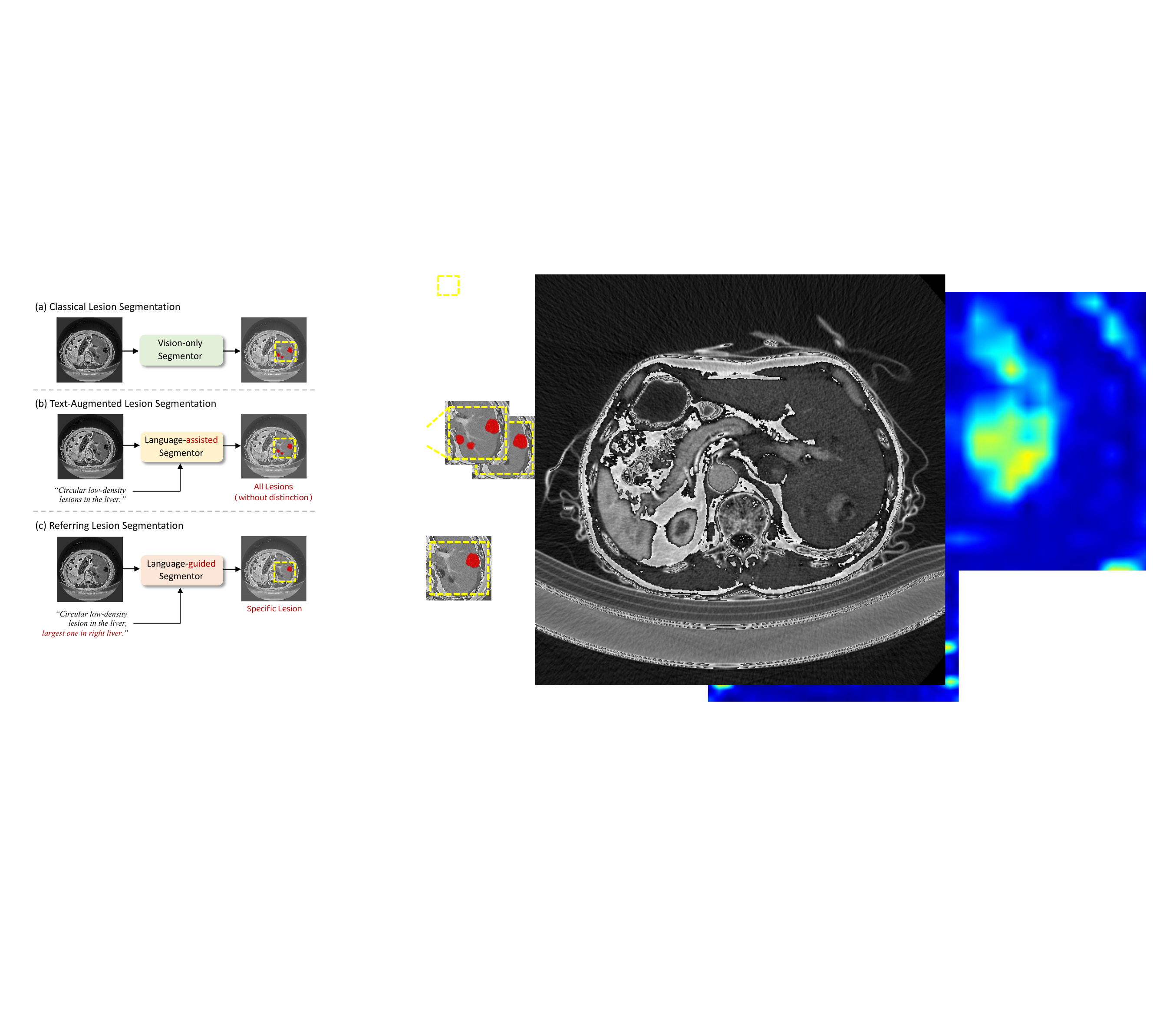}
  % \vspace{-10pt}
  \caption{Comparison between Referring Lesion Segmentation  and conventional lesion segmentation tasks.
  % , illustrating the distinction in feature handling process between the proposed LSMS and existing medical segmentation methods.
  % the distinct approach of the proposed  LSMS in contrast to existing medical segmentation methods.
  In the images displaying segmentation results, the regions highlighted in red represent the Ground Truth.
  % The key region within the image have been enlarged, with a yellow rectangular box serving as a reference for location.
  % To facilitate intuitive correspondence with the left and right orientations mentioned in the text for ease of observation, the CT images presented in this paper have been mirrored.
  For intuitive correspondence with the left-right references in the text, all CT images in this paper have been mirrored horizontally.}
  % \Description{A woman and a girl in white dresses sit in an open car.}
  \label{fig1} 
  % \vspace{-15pt}
\end{figure}

\IEEEPARstart{T}{he}  significance of  lesion segmentation in medical image analysis has been widely  recognized \cite{ronneberger2015u,zhou2019unet++}.
It enables the precise identification and delineation of pathological regions, which is essential for accurate diagnosis and treatment planning.
As shown in Fig.~\ref{fig1}(a), the \emph{Classical Lesion Segmentation} task involves inputting a medical image and obtaining segmentation results encompassing {lesions} within the image \cite{chen2021transunet,wang2022uctransnet}. 
With the advancement of multi-modal research, studies have emerged that incorporates textual input as supplementary information for segmentation.
% that supplement image segmentation with textual input.
Fig.~\ref{fig1}(b) illustrates \emph{Text-Augmented Lesion Segmentation}, wherein diagnostic texts provided by physicians aid in image interpretation \cite{li2023lvit}. This task involves inputting a medical image along with its  diagnostic text and outputting segmentation results for  
% \textcolor[rgb]{0.8,0,0}{{All Lesions}} 
{All Lesions} without distinction. 
% present \cite{li2023lvit}. 
However, in clinical practice, physicians often need to segment specific lesions to provide tailored diagnosis and treatment, thus rendering \emph{Conventional Lesion Segmentation}
\footnote{In this paper, \emph{Conventional Lesion Segmentation} is used as a collective term to refer to the two tasks illustrated in Figure 1: (a) Classical Lesion Segmentation and (b) Text-Augmented Lesion Segmentation. \\ \\} 
tasks inadequate for practical applications. For instance, in radiation therapy, the accuracy of tumor segmentation directly influences the precision of radiation beam targeting, which determines the effectiveness of the treatment and the potential risk of side effects. 
% For instance, in radiation therapy, the accuracy of specific tumor segmentation directly influences the precision of radiation beam targeting, which determines treatment effectiveness and the risk of side effects.
As shown in Fig.~\ref{fig1}(a) and (b), \emph{Conventional Lesion Segmentation} tasks treat all lesions equally, which is insufficient for clinical scenarios.
% However, in clinical practice, physicians often need to segment \textbf{specific lesions} to aid in diagnosis and treatment, thus rendering \emph{Conventional Lesion Segmentation}
% \footnote{In this paper, \emph{Conventional Lesion Segmentation} is used as a collective term to refer to the two tasks illustrated in Figure 1: (a) Classical Lesion Segmentation and (b) Text-Augmented Lesion Segmentation.\\ \\} 
% tasks inadequate for practical applications. 
Text, as a convenient medium for expressing physicians' needs, can be used to indicate target segmentation objects.
% In clinical practice, physicians often  require the segmentation of specific lesions to assist in diagnosis and treatment, thus rendering conventional image segmentation methods inadequate for practical applications. 
% Language, serving as a convenient medium for expressing physician requirements, can be employed as an auxiliary modality for input. 
Therefore, we introduce a new task of  \emph{Referring Lesion Segmentation (RLS)}, 
% where the input comprises medical images along with referring language expression pointing to specific lesions within the images, with the anticipated output being the segmentation results corresponding to the lesions indicated in the reference expression.
where medical images are accompanied by a language expression that indicate a 
% \textcolor[rgb]{0.8,0,0}{{Specific Lesion}}
{Specific Lesion}
% \colorbox[rgb]{0,0,0}{Specific Lesion}
within the image. 
% The objective is to generate segmentation results corresponding to the lesions mentioned in the reference language expressions.
Fig.~\ref{fig1}(c) represents the introduced task, RLS, which involves segmenting the largest lesion in the right liver as described by the language expression.
% , whereas the task in Fig.~\ref{fig1}(b) does not distinguish between multiple lesions.
Compared to the task in Fig.~\ref{fig1}(b), RLS emphasizes the ability to differentiate  lesions based on the text.
To support  evaluation of RLS, we developed  Reference Hepatic Lesion Segmentation (RefHL-Seg) dataset—the first benchmark specifically designed for this new task.

% 分析现有方法
% Deep learning methods have demonstrated exceptional performance in image segmentation tasks. In recent years, research on vision-language models has also rapidly progressed, exhibiting superior performance in natural image tasks.

\begin{figure}[t!]
  \centering
  \includegraphics[width=\linewidth]{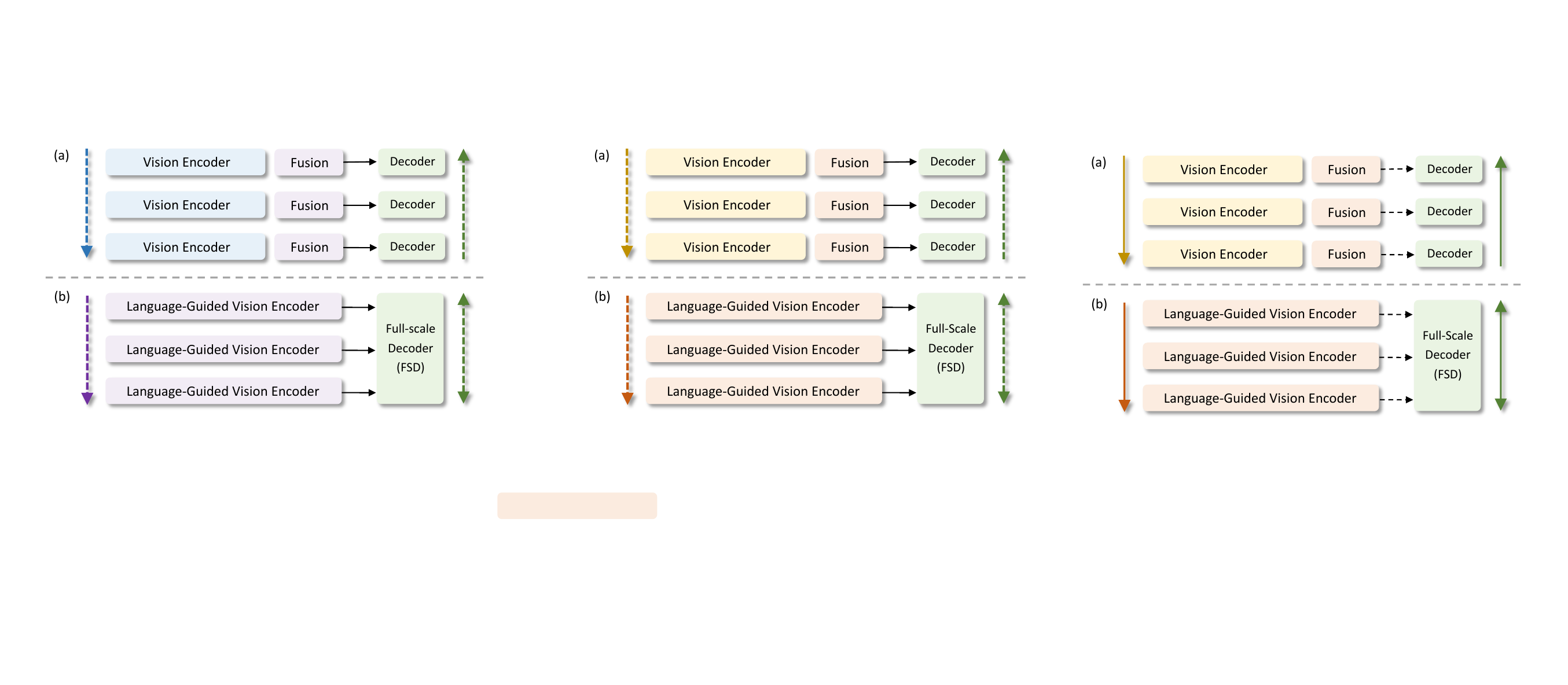}
  % \vspace{-10pt}
  \caption{Comparison of Transformer-based architectures. (a) Existing architectures  for related tasks. (b) our LSMS for RLS.}
  % \Description{A woman and a girl in white dresses sit in an open car.}
  \label{fig8} 
  \vspace{-10pt}
\end{figure}

Deep learning methods have demonstrated outstanding performance in medical image segmentation tasks. 
% in classical medical image segmentation, methods based on the U-Net \cite{ronneberger2015u} architecture are widely utilized \cite{oktay2018attention,cao2022swin}. 
\emph{Classical Lesion Segmentation} methods often relies on architectures such as U-Net \cite{ronneberger2015u} or Transformer \cite{vaswani2017attention}. Methods combining these two architectures generally down-sample feature maps to acquire high-level contextual information, followed by up-sampling to reconstruct spatial dimensions~\cite{chen2021transunet,cao2022swin}. 
% These methods leverage the skip connections and upsampling paths within the structure to fully utilize visual information, leading to superior performance. 
To further enhance segmentation performance, approaches incorporating language modality have emerged \cite{li2023lvit,huang2024cross}. 
% As shown in Fig.~\ref{fig5}, these transformer-based methods include a fusion module at the end of each stage to fuse extracted visual features with linguistic  information.
% In tasks related to RLS,  Transformer-based designs leverage long-range dependencies and hierarchical structures to enhance performance.
Recently, Transformer-based models have shown great promise in this area, as their ability to model long-range dependencies facilitates  effective integration of visual and linguistic information.
% , although there is still potential for further refinement. 
% Typically, LAVT \cite{yang2022lavt} is a mature   \emph{Natural Image Referring Segmentation} model, while LViT~\cite{li2023lvit} is the latest  model for \emph{Text-Augmented Lesion Segmentation}.
% These typical approaches are designed around the Transformer architecture, 
As shown in Fig.~\ref{fig8}(a), existing Transformer-based methods include a fusion module at the end of each stage to integrate the extracted visual features with linguistic information.
Models utilizing this architecture include the mature \emph{Natural Image Referring Segmentation} model LAVT \cite{yang2022lavt} and the recent \emph{Text-Augmented Lesion Segmentation} model LViT \cite{li2023lvit}.
% Specifically, current approaches treat visual feature extraction and cross-modal fusion as two separate steps, leaving room for improvement in  visual-linguistic alignment within the semantic space.
By introducing a multi-scale structure, they effectively exploit rich vision-language knowledge. 
However, these methods are not directly applicable to the RLS task.
Existing methods treat visual feature extraction and cross-modal fusion as two independent steps, leaving room for improvement in achieving visual-linguistic alignment within the semantic space.
% fuse visual and linguistic features during the encoding stage of the model, aiding in the acquisition of additional contextual information.
% The latest referring segmentation methods in natural images also utilize the similar architecture \cite{yang2022lavt,ouyang2023slvit}.
% and further improving segmentation performance. 
% However, existing methods employing sequential structures result in single-scale representations at each level, which may not adequately accommodate the size and shape variations inherent in medical images.
% However, due to the varying size and shape inherent in medical images, the fusion of single-scale representations at each stage with linguistic features alone is insufficient for precise object localization.
During decoding, they adopt sequential structures that tend to produce single-scale representations at each level, while enhancing inter-scale interaction may help  capture core semantic information.
% However, due to the inherent variability in the size and shape of objects in medical images, the use of a single-scale representation combined with linguistic features at each stage is insufficient for precise object localization.
% % in complex visual environments. 
% Moreover, the sequential structure adopted in the decoding stage lacks a comprehensive understanding of features at various scales, thereby constraining  segmentation performance.

% Upon analyzing the requirements of the RLS task, and previous successful endeavors, we contend that an effective approach should possess the following features:
% (i) A strong visual-language encoder to acquire detailed knowledge from the visual environment and engage in close interaction with linguistic information. Enriched local visual features, under the guidance of language information, can aid the model in precise object localization.
% (ii) Comprehensive multi-scale interactions to globally model the complex differences between scales and obtain the most effective global visual-linguistic features. Given the complexity of the visual environment in medical imaging, learning complementary information between scales can assist in accurately identifying lesion edges during segmentation.
% Therefore, considering the aforementioned analysis, we propose a RLS architecture (see Fig.~\ref{fig1}(c)), namely Language-guided Scale-aware MedSegmentor (LSMS).

Upon analyzing the requirements of RLS and previous efforts in related tasks, we identify two key aspects that address the unique challenges of RLS while aligning with conventional segmentation goals:
(i) Robust vision-language modeling. 
% The substantial variations in size and shape between objects in the medical visual environment pose a challenge for integrating visual and linguistic features effectively. 
In the medical visual environment, the notable variations in size and shape among objects make it crucial to effectively model visual-linguistic consistency for accurate object localization.
Fusing visual and linguistic features after visual feature extraction may overlook the rich local visual information correlated with linguistic guidance, thereby impacting the model's object localization performance.
(ii) Comprehensive multi-scale interaction. Globally modeling the complex differences across scales enables the extraction of optimal global visual-linguistic features. Given the complexity of the visual environment in medical images, neglecting complementary information between scales may result in insufficient capability to identify lesion boundaries during segmentation.

\begin{figure*}[ht!]
    \centering
    \includegraphics[width=0.97\linewidth]{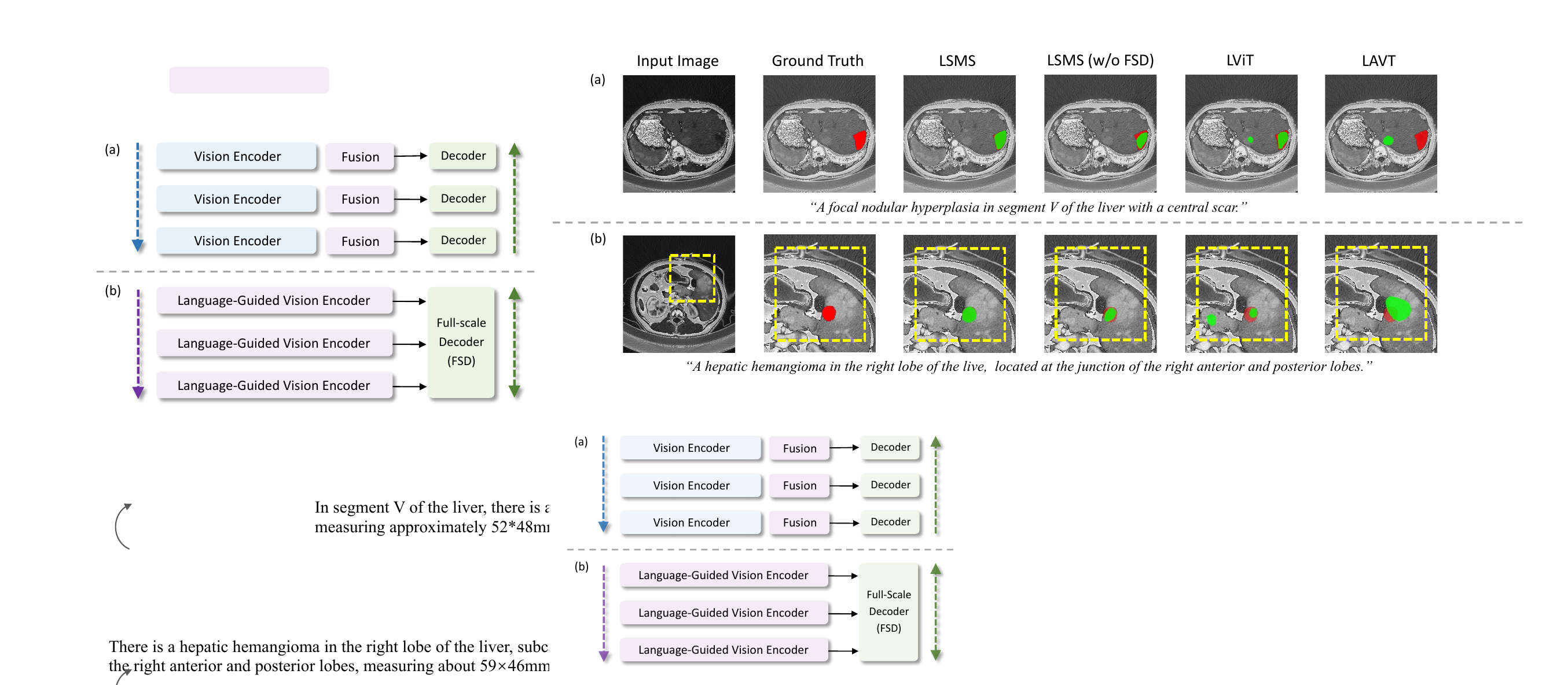}
    % \vspace{-15pt}
    \caption{
    Qualitative results of different approaches.
    The red regions denote the Ground Truth, while the green regions represent the segmentation results of our LSMS, LSMS (w/o FSD), LViT \cite{li2023lvit} and LAVT \cite{yang2022lavt}. In sample (b), for ease of observation, the key region within the image have been enlarged, with a yellow rectangular box serving as a reference for location.
    % The reference expressions provided for inputs (a) and (b) are "In segment V of the liver, a focal nodular hyperplasia with a central scar, measuring approximately $52\times48$ mm" and "A hepatic hemangioma in the right lobe of the liver, subcapsularly located at the junction of the right anterior and posterior lobes, measuring about $16\times17$ mm"
    }
    \label{fig5}
    \vspace{-5pt}
\end{figure*}

% To address these challenges, we introduce a novel RLS architecture shown in Fig.~\ref{fig5}(b), namely \textbf{L}anguage-guided \textbf{S}cale-aware \textbf{M}ed\textbf{S}egmentor (LSMS).
% To address these challenges, 
In light of the aforementioned analysis,
we propose a  model named \textbf{L}anguage-guided \textbf{S}cale-aware \textbf{M}ed\textbf{S}egmentor (LSMS), as illustrated in Fig.~\ref{fig8}(b). 
Within LSMS, we introduce a Scale-aware Vision-Language Attention (SVLA) module embedded in the encoder block. 
% SVLA facilitates the acquisition of scale-aware visual knowledge and fosters close interactions with linguistic features, thereby significantly improving visual-linguistic consistency.
% SVLA enhances the acquisition of scale-aware visual knowledge by employing convolutional kernels of various sizes, facilitating close interaction with linguistic features and thereby  improving visual-linguistic consistency.
% SVLA concurrently performs visual feature extraction and vision-language alignment, facilitating close interaction  and thereby enhancing the consistency between visual and linguistic representations.
SVLA captures scale-aware visual knowledge and models visual-linguistic relationships in an integrated manner, enhancing the visual-linguistic alignment  in the semantic space.
As shown in Fig.~\ref{fig5}, LSMS (w/o FSD) equipped with SVLA exhibits a significant enhancement in lesion localization capability compared to the results of existing vision-language model LViT \cite{li2023lvit} and LAVT \cite{yang2022lavt}.
% LViT and LAVT are well-established vision-language segmentation models in the domains of medical and natural image segmentation, respectively.
Additionally, we devise a Full-Scale Decoder (FSD) 
that globally models multi-modal information by aligning and integrating multi-modal feature maps from various scales, thereby enhancing the comprehension of details within complex medical images.
% for globally modeling multi-modal information from different scales, enhancing the understanding of details in complex medical visual environment. 
In Fig.~\ref{fig5}(b), LSMS, when contrasted with LSMS (w/o FSD), exhibits a more precise prediction of lesion boundaries during segmentation.

Additionally, we have designed a specialized loss function to constrain the model’s training, which includes the Segmentation Loss $\mathcal{L}_{Seg}$ for the segmentation results and the Vision-Language Contrastive Loss $\mathcal{L}_{Con}$ for the linguistic features and final multi-modal features. Specifically, $\mathcal{L}_{Seg}$ enhances the focus on the object boundaries, thereby improving edge detection capabilities. Meanwhile, $\mathcal{L}_{Con}$ further aligns the visual knowledge of the target region with the linguistic information, leading to improved accuracy in target localization.

In summary, our contributions encompass four aspects:
\begin{enumerate} 
% [itemsep=2pt,topsep=0pt,parsep=0pt]

% \item We introduce a Scale-aware Vision-Language Attention module tailored for visual-linguistic fusion, enabling the acquisition of rich visual knowledge and facilitating close interaction with linguistic features. This enhances visual-linguistic consistency and improves the model's object localization capability.

% \item We propose a full-scale decoder that globally models multi-modal features from different stages, facilitating a comprehensive understanding of the visual environment with linguistic cues. This aids in the more accurate  prediction of lesion boundaries during segmentation.

\item We introduce a new task of RLS, which entails segmenting the target object in medical images based on the language expression. 
% The application of RLS in clinical practice can  enhance diagnostic efficiency for physicians.
% This task is designed to address specific requirements in clinical practice. 
% We introduce a novel task for medical image object localization and segmentation based on reference expressions, termed RLS, which is tailored to address specific needs in clinical practice.
% We have developed the RefHL-Seg dataset specifically for the RLS task to facilitate its training and evaluation.
We have established the RefHL-Seg dataset as a new benchmark for RLS.
% , providing a standardized foundation for training and evaluation.

% \item We curate the Reference Hepatic Lesion Segmentation (RefHL-Seg)  dataset comprising  {abdominal CT images}, {text annotations} specifying particular lesions, and corresponding {segmentation ground truth}. This dataset serves for training and evaluating the RLS task.

% \item We develop the RefHL-Seg dataset for training and validating the RLS task. RefHL-Seg consists of 2,283 abdominal CT slices from 231 cases.
% , including abdominal CT images, textual annotations specifying particular lesions, and corresponding segmentation ground truths.

\item 
We propose LSMS for lesion segmentation, comprising a SVLA module to improve object localization capability and a full-scale decoder to enhance the accuracy of lesion boundary prediction.
% during segmentation.
% We propose the LSMS model for RLS, incorporating a SVLA module to optimize cross-granularity alignment and enhance target localization performance.

\item 
% We design a custom loss function to improve boundary delineation accuracy, facilitating more precise segmentation of target objects.
We design a specialized loss function to optimize fine-grained discrimination and visual-linguistic alignment, thereby enhancing the accuracy.

\item We conduct comprehensive experiments on the RefHL-Seg dataset for RLS, as well as on datasets for conventional lesion segmentation tasks. Experimental results demonstrate the superiority of LSMS over current state-of-the-art methods with lower computational costs.

\end{enumerate}

The preliminary work was presented as a conference paper at the International Joint Conference on Artificial Intelligence (IJCAI) 2023 \cite{ouyang2023slvit}. This paper extends the prior work by introducing a new task (RLS), a  dataset (RefHL-Seg) as the new benchmark, as well as method and experimental enhancements. We optimize the scale-wise module design in the model and propose a specialized loss function, which improves the model's performance in the domain of lesion segmentation. Furthermore, we apply the proposed model to the newly introduced RLS task and validate it on conventional lesion segmentation benchmarks, demonstrating the generalizability  of our approach.

\section{Related Works}

\subsection{Medical Image Segmentation}

For medical image segmentation tasks, early methods \cite{long2015fully,chen2017deeplab} extended from image classification networks to achieve semantic segmentation, among which the Fully Convolutional Network (FCN) \cite{long2015fully} is an end-to-end semantic segmentation network [18]. Multi-scale architectures have been proven to enhance segmentation performance based on CNN methods. U-Net \cite{ronneberger2015u} is considered a pioneer in medical image segmentation. U-Net++ \cite{zhou2019unet++} further improved U-Net by enhancing skip connections to bridge the semantic gap between shallow encoder layers and deep decoder layers. MS-DualGuided \cite{sinha2020multi} focused on both spatial and channel dimensions of feature maps at different scales. CA-Net \cite{wang2022multimodal} proposed a multi-scale context-aware network for multi-modal medical image segmentation.  Subsequently, the introduction of Transformer \cite{vaswani2017attention} was found to enhance medical image segmentation performance.
TransUNet \cite{chen2021transunet} integrated the power of Transformers with the U-Net architecture for improved instance segmentation. Swin-Unet \cite{cao2022swin} specifically designed to leverage the hierarchical features and shifted window mechanisms of Swin Transformer \cite{liu2021swin}, thereby enhancing the model's capability to capture both local and global context.
Furthermore, LViT \cite{li2023lvit} proposed a medical image segmentation model with textual assistance, introducing text input as auxiliary information during the encoding stage of segmentation. RecLMIS \cite{huang2024cross} proposes a novel cross-modal conditioned method for {Text-Augmented Lesion Segmentation}. 
% These methods acquire feature maps through progressively down-sampling, yielding low-resolution representations. And they utilize a sequential up-sampling structure in the decoding stage, lacking a global understanding of multi-scale knowledge.
While current research incorporates text as an aid for image understanding, existing benchmarks and methods are unable to  distinguish specific objects in medical images, thereby falling short of fully addressing the practical needs of physicians.

\begin{figure*}[ht!]
    \centering
    \includegraphics[width=1.0\linewidth]{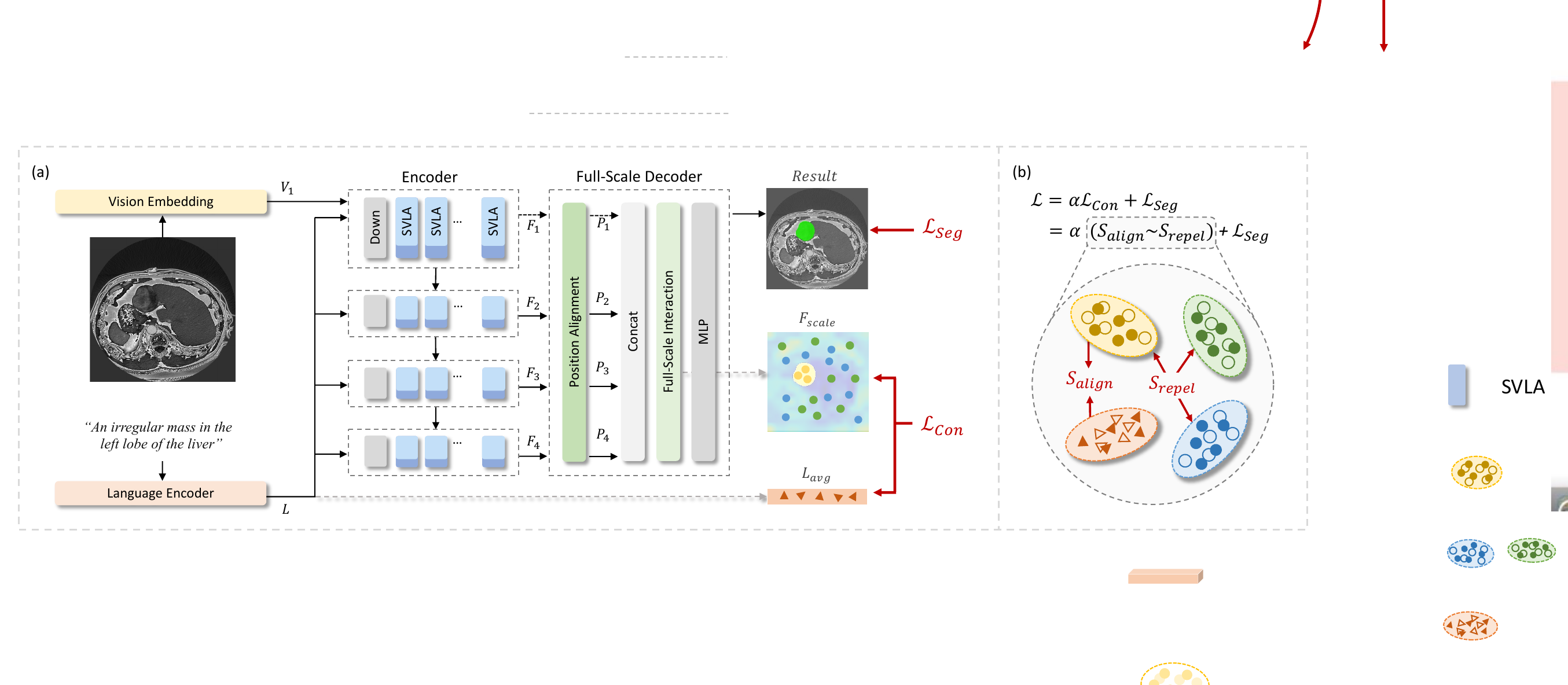}
    \caption{(a) An illustration of LSMS.
    Initially, the input image and the reference expression are embedded separately through the vision embedding block and the BERT \cite{devlin2018bert} language encoder, yielding visual feature $V_1$ and linguistic feature $L$, which are then fed into the language-guided vision encoder. The encoder incorporates the Scale-aware Vision-Language Attention (SVLA) module to interact between visual knowledge from different receptive fields and linguistic features. The encoder blocks at each stage learn rich multi-modal features $F_i, i \in \{ 1,2,3,4\}$, which are subsequently fed into the full-scale decoder. Through the Position Alignment block, $F_i, i \in \{ 1,2,3,4\}$ uniformly resize the feature maps of various scales while preserving channel disparities, resulting in $P_i, i \in \{ 1,2,3,4\}$. Features $P_i, i \in \{ 2,3,4\}$ are then globally modeled across scales for final segmentation.
    (b) An illustration of the operational mechanism of the vision-language contrastive loss $\mathcal{L}_{Con}$. 
    % The figure demonstrates the alignment (attraction) of positive pixel features with the average linguistic feature $L_{avg}$, while simultaneously repelling the features of negative pixels, thereby enhancing the separation between target and non-target regions in the feature space.
    }
    \label{fig2}
    \vspace{-5pt}
\end{figure*}

% \subsection{Vision-Language Segmentation Models}
\subsection{Natural Image Referring Segmentation}
Early referring segmentation methods in natural images \cite{hu2016segmentation,liu2017recurrent,li2018referring} typically combined visual and linguistic features by concatenation, utilizing FCNs for cross-modal feature learning and prediction. In contrast, attention-based methods, such as vision-guided linguistic attention \cite{shi2018key} and Cross-Modal Self-Attention \cite{ye2019cross}, focus on aligning visual content with linguistic information. Bi-directional relationship networks \cite{hu2020bi} capture mutual guidance, while others \cite{yu2018mattnet,huang2020referring} use sentence structure to model cross-modal attributes. 
% \cite{hui2020linguistic} guides context aggregation using word-level syntactic structures. 
% Recently, Transformer-based methods have significantly improved long-distance cross-modal dependency modeling in referring segmentation.
More recently, Transformer-based methods have enhanced the modeling of long-range cross-modal dependencies, leading to significant advancements in referring segmentation.
% The capability of vision-language models to encode rich multi-modal representations has played a crucial role in the field of computer vision \cite{wang2017multi,zhao2021m3tr}. ViLT \cite{kim2021vilt}, by simplifying the processing of visual inputs in CLIP \cite{radford2021learning}, proposed a more parameter-efficient architecture. This architecture enables interaction layers to handle visual features effectively without relying on separate deep visual embeddings. Subsequently, there have been significant research efforts in image segmentation \cite{li2021gt}, where text information is utilized to enhance model segmentation capabilities. 
VLT \cite{ding2021vision} and EFN \cite{feng2021encoder} utilize a Transformer-based encoder-decoder framework, employing attention mechanisms in decoding stages to augment contextual information. LAVT \cite{yang2022lavt} adopt an early fusion approach, modeling multi-modal context within the Transformer encoders. 
Prompt-RIS \cite{shang2024prompt}  leverages bidirectional prompting for better vision-language interaction.
% Some methods \cite{bhalodia2021improving,muller2022joint} have started using text information to assist in medical image analysis. 
% \cite{qin2022medical} leveraged the knowledge transfer ability of pretrained vision-language models \cite{li2022grounded} to the medical domain by employing well-designed medical prompts.
% LViT \cite{li2023lvit} introduced textual assistance as auxiliary information during the encoding stage for segmentation. RecLMIS \cite{huang2024cross} proposes a novel cross-modal conditioned method. 
% These methods utilize a sequential up-sampling structure in the decoding stage, lacking a global understanding of multi-scale knowledge.
% Due to the complexity of medical visual environment, existing methods for \emph{Natural Image Referring Segmentation} struggle to achieve explicit alignment between text and images in RLS. To address this, we propose a design involving close interaction between scale-aware visual knowledge and linguistic features to facilitate modeling of visual-linguistic relationships.
Current methods for {Natural Image Referring Segmentation}  treat visual feature extraction and cross-modal fusion as two separate steps, and use sequential upsampling structures during  decoding. However, due to the complexity of medical visual environments, these methods struggle to meet the  requirements of RLS. In this paper, we propose  the SVLA, which models visual information and visual-linguistic relationships in an integrated manner, and employ a full-scale decoder to promote comprehensive understanding of multi-scale knowledge.

% \vspace{15pt}

\section{Language-Guided Scale-Aware Medical Segmentor}

\subsection{Overview}

In the proposed Language-guided Scale-aware MedSegmentor (LSMS), we learn visual knowledge from diverse receptive fields and tightly engage with linguistic features. Subsequently, leveraging a full-scale decoder, we globally model multi-modal information across all scales, thereby facilitating  RLS task. The training loss includes the Segmentation Loss \(\mathcal{L}_{{Seg}}\)  and the vision-language contrastive loss \(\mathcal{L}_{{Con}}\).
The overall architecture of LSMS is presented
in Fig.~\ref{fig2}.

Given an image and a language expression, our model predicts the segmentation mask corresponding to the language reference. Our LSMS follows a workflow composed of \textbf{\emph{[language-guided vision encoder]  - [full-scale decoder]}}. LSMS comprises four stages, each with varying numbers of encoder blocks and different feature map resolutions. The encoder (Sec.~III.B) incorporates a novel Scale-aware Vision-Language Attention module (Sec.~III.C), which learns rich visual knowledge through convolutions of different sizes and closely interacts with linguistic features. Additionally, we propose a full-scale decoder (Sec.~III.D) to globally model multi-modal information across multiple scales, enhancing the understanding of context details. 
During training, we employ the  \(\mathcal{L}_{{Seg}}\) to supervise the segmentation results and the \(\mathcal{L}_{{Con}}\) to constrain the linguistic features and the final multi-modal representations (Sec.~III.E).
In the following subsections, we elaborate on each component of LSMS.

\subsection{Language-Guided Vision Encoder}

To facilitate deep interaction between linguistic features and the complex visual environment of medical images, our encoder leverages a novel attention module   (Sec.~III.C) to perceive visual details in the image under linguistic guidance, thus obtaining valuable multi-modal representations. 
The structure of the encoder block is depicted in Fig.~\ref{fig3}(a).

As illustrated in Fig.~\ref{fig2}, our encoding phase comprises four stages with decreasing feature map resolutions. The $i$-th stage consists of $N_i$ encoder blocks. For the sake of clarity, we assume that $N_i=1, i \in \{1,2,3,4\}$ in this section.
LSMS receives an input image alongside a reference expression. We extract the linguistic feature $L \in \mathbb{R}^{C_{l} \times T}$ using the  BERT \cite{devlin2018bert} language encoder, where $T$ denotes the number of words, and $C_l$ represents the channel number of the  linguistic feature. Similarly, the input image undergoes an embedding block to yield the initial visual input $V_1 \in \mathbb{R}^{C_{v1} \times H_1 \times W_1}$ for the encoder, where $H_1$ and $W_1$ represent the height and width of the  visual feature, and $C_{v1}$ represents the channel number.
Each stage comprises a down-sampling block and a stack of encoder blocks. For each stage, the aggregation of the multi-modal feature maps $F_i  \in \mathbb{R}^{C_{vi} \times H_i \times W_i}$ can be expressed as:
\begin{equation}
    F_i = \begin{cases}
\mathit{LGVE}(V_1, L), & i=1 \\
\mathit{LGVE}(Down(F_{i-1}), L),& i=2,3,4 \\
\end{cases}
\end{equation}
% \begin{equation}
%     F_i =
%     \begin{cases}
%     \begin{alignat}{1}
%     & LGVE(V_1, L), &  \text{if } i &= 1, \\
%     & LGVE(Down(F_{i-1}), L), & \text{if } i &= 2,3,4.
%     \end{alignat}
%     \end{cases}
% \end{equation}
\noindent where $i$ denotes the index of the stage, the function $Down(\cdot)$ represents the down-sampling block, and $\mathit{LGVE}(\cdot)$ denotes the encoder block. The visual input provided to the encoder block is obtained by $V_i=Down (F_{i-1})$. The down-sampling block consists of a convolutional layer with a stride of 2 and a kernel size of 3 × 3, followed by batch normalization. 

As depicted in Fig.~\ref{fig3}(a), the architecture of encoder blocks follows the design of ViT \cite{dosovitskiy2020image}, but we introduce a SVLA module (Sec.~III.C) to replace the conventional self-attention mechanism. The workflow of the encoder block is illustrated by the following:
\begin{equation}
    F_i^{'} = \mathit{Norm}(\mathit{SVLA}(V_i,L)+V_i),
\end{equation}
\begin{equation}
    F_i = \mathit{Norm}(FeedForward(F_i^{'})+F_i^{'}),
\end{equation}

\noindent where $\mathit{Norm}(\cdot)$ and $FeedForward(\cdot)$ represent normalization and feed forward layers, $\mathit{SVLA}(\cdot)$ denotes SVLA module, and the output of $\mathit{SVLA}(V_i,L)$ is labeled as $F_i^{Att}$.

\begin{table}[]
\label{table2}
\large
% \scriptsize
% \footnotesize
\centering
\caption{Detailed settings of different stages in our LSMS.}
\resizebox{0.48\textwidth}{!}{
\renewcommand\arraystretch{1.4}

\begin{tabular}{c|cccc}
\toprule [0.6pt]

            & Stage 1              & Stage 2              & Stage 3              & Stage 4              \\ 
\midrule [0.4pt]
% output size &                      &                      &                      &                      \\
number of blocks ($N_i$)     & 3 & 3 & 5 & 2 \\
\midrule [0.4pt]
visual output size       &    $\frac{H}{4} \times \frac{W}{4}$    &  $\frac{H}{8} \times \frac{W}{8}$   &  $\frac{H}{16} \times \frac{W}{16}$    &   $\frac{H}{32} \times \frac{W}{32}$   \\
\midrule [0.4pt]
channel      &  64       &     128         &     320      &        512   \\

\bottomrule [0.6pt]
\end{tabular}
}
\label{tab1}
\vspace{-5pt}
\end{table}

\subsection{Scale-aware Vision-Language Attention}

In medical images, instances vary greatly in size and shape, posing a challenge in pinpointing lesions referred to  linguistic cues within intricate visual contexts. Addressing this, 
% we integrate linguistic features with visual knowledge across multiple scales by employing convolutional kernels of varying sizes to capture visual information from different receptive fields.
we learn scale-aware visual knowledge  from diverse receptive fields by employing convolutional kernels of varying sizes, integrating linguistic features with visual knowledge across multiple scales.

\begin{figure}[t!]
    \centering
    \includegraphics[width=\linewidth]{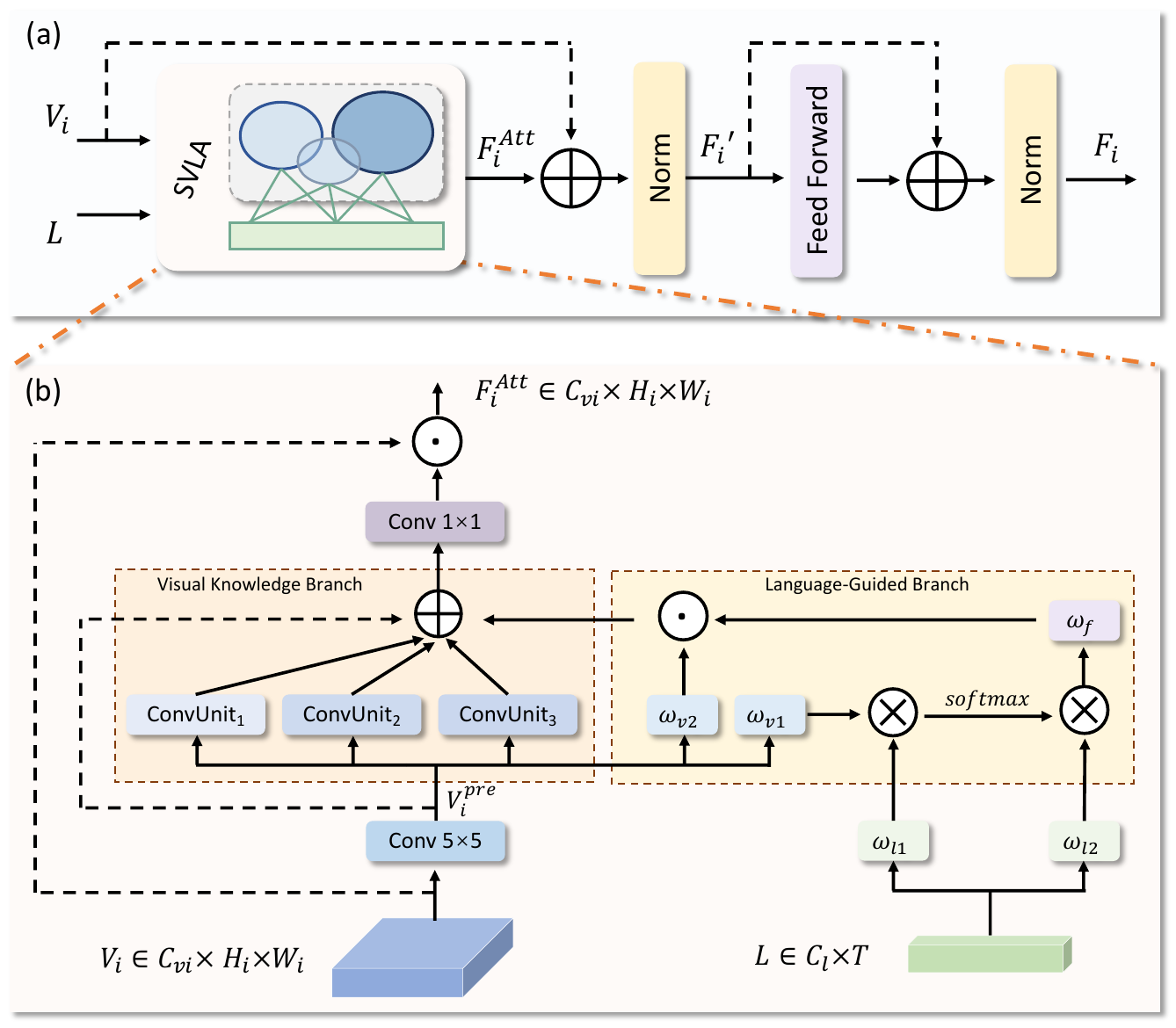}
    % \vspace{-8pt} 
    \caption{ (a) An illustration of the encoder block in the Language-Guided Vision Encoder. (b) An illustration of the Scale-aware Vision-Language Attention module.
    }
    \label{fig3}
    \vspace{-5pt} 
\end{figure}

As depicted in Fig.~\ref{fig3}(b), our proposed attention mechanism, termed Scale-aware Vision-Language Attention (SVLA), initially captures preliminary visual features through a convolution operation, and then employs Visual Knowledge Branch and  Language-Guided Branch to model rich visual knowledge and visual-linguistic relationships, followed by a $1 \times 1$ convolution operation to learn the interplay between the branches.
In $i$-th stage, with visual input $V_i \in \mathbb{R}^{C_{vi} \times H_i \times W_i}$ and language input $L \in \mathbb{R}^{C_l \times T}$, we obtain the preliminary visual feature map $V_i^{pre} \in \mathbb{R}^{C_{vi} \times H_i \times W_i}$  by the formula $V_i^{pre} = Conv_{5 \times 5}(V_i)$.

\paragraph{\textbf{Visual Knowledge Branch}}
To accommodate the characteristics of medical image visual environment, we devised ConvUnits to capture scale-aware visual knowledge. ConvUnits comprises diverse convolution kernels, with each unit consisting of  a $1 \times d_j$ and a $d_j \times 1$ convolution operations, where $j \in \{1,2,3\}$. Each SVLA module comprises three ConvUnits, aimed at capturing scale-aware visual knowledge from various receptive fields. Visual knowledge $Att_i^V  \in \mathbb{R}^{C_{vi} \times H_i \times W_i}$ can be derived using the following formula:
\begin{equation}
    Att_i^{V} = \sum_{j=1}^3 ConvUnit_j(V_{i}^{pre}),
\end{equation}%

\noindent where $ConvUnit_j(\cdot)$ indicates $j$-th ConvUnit. 
The rectangular convolution kernels in ConvUnits enable the acquisition of  detailed visual information with low computational costs.

\paragraph{\textbf{Language-Guided Branch}}

We employ the Language-Guided Branch to model relationships between linguistic information and various visual coordinates, facilitating the guidance of lesion localization in the complex visual environment. The steps to obtain language-guided knowledge $Att_i^{L} \in \mathbb{R}^{C_{vi} \times H_i \times W_i}$ are as follows:
\begin{equation}
    V_{i1},V_{i2} = flatten(\omega_{v1}(V_{i}^{pre}) ), \omega_{v2}(V_{i}^{pre}) ,
\end{equation}%
\begin{equation}
    L_{i1},L_{i2} = \omega_{l1}(L), \omega_{l2}(L),
\end{equation}%
\begin{equation}
    \alpha_i = V_{i1}^\top L_{i1},
\end{equation}%
\begin{equation}
    Att_i^{L\prime} = unflatten((softmax(\frac{\alpha_i}{\sqrt{C_l}})L_{i2}^\top)^\top), 
\end{equation}%
\begin{equation}
    Att_i^{L} = \omega_{f}(Att_i^{L\prime}) \odot V_{i2}, 
\end{equation}%
\noindent where $\omega_{v1}$, $\omega_{v2}$, $\omega_{l1}$, $\omega_{l2}$, $\omega_{f}$ are projection functions, $flatten(\cdot)$ denotes the operation of flattening the two spatial dimensions into a single dimension along the rows, $unflatten(\cdot)$ indicates the opposite operation of $flatten(\cdot)$, and $\odot$ is element-wise matrix multiplication operation.  $\omega_{l1}$ and $\omega_{l2}$ each is implemented as a 1×1 convolution, yielding channels of size $C_{vi}$. $\omega_{v1}$, $\omega_{v2}$ and $\omega_{f}$ each is defined as a 1×1 convolution and an instance normalization.

\paragraph{\textbf{Comprehensive Attention}}

We integrate information from the Visual Knowledge Branch and the Language-Guided Branch to compute comprehensive attention weights through convolution, thereby reweighting the input $V_i$ to the SVLA module. The feature $F_{i}^{Att}  \in \mathbb{R}^{C_{vi} \times H_i \times W_i}$ is obtained using the following:
\begin{equation}
    F_{i}^{Att} = Conv_{1 \times 1}(V_i^{pre} + Att_i^{V} + Att_i^{L} ) \odot V_{i},
\end{equation}%

\noindent where $Conv_{1 \times 1}$ represents the $1 \times 1$ convolution operation,
and $\odot$ is element-wise matrix multiplication operation.

\subsection{Full-Scale Decoder}

To accommodate the complexity of medical visual environment, 
% In order to integrate linguistic cues for comprehending complex medical visual environment,
a comprehensive cross-scale understanding of visual-linguistic contexts is necessary. Therefore, we devised a Full-Scale Decoder (FSD) to capture high-level semantics following the encoder. Unlike previous methods \cite{zhao2017pyramid,xie2021segformer,fu2019dual} with sequential structures, we globally process multi-modal features of various scales after aligning their positions. We employ Position Alignment to map them to the same feature map size of $F_1$ while retaining their original channel numbers. Subsequently, we concatenate features from different scales, pass them through a lightweight InterScale block for globally modeling multi-scale contexts, and finally generate segmentation predictions. The process is as follows:
\begin{equation}
    P_1, P_2,P_3,P_4 = PositionAlign(F_1, F_2,F_3,F_4) ,
\end{equation}%
\begin{equation}
    Out = Seg(InterScale(Concat[P_2,P_3,P_4])) ,
\end{equation}%

\noindent where $PositionAlign(\cdot)$ represents the  Position Alignment operation, $InterScale(\cdot)$ is implemented as a lightweight Hamburger \cite{geng2021attention} function, and $Seg(\cdot)$ indicates a $1 \times 1$ convolution and an up-sampling function for final prediction. $PositionAlign(\cdot)$ is realized through bilinear interpolation operations.

It is noteworthy that we exclusively utilize features generated from $2,3,4$-th stages for global decoding. This choice is informed by the observation that the shallow features from the first stage exhibit a lower degree of visual-linguistic consistency, encompassing redundant information from irrelevant regions in the images. 
% This redundancy impedes lesion localization based on linguistic cues and hinders segmentation. 
% This redundancy hampers lesion localization  and hinders segmentation.
This redundancy hampers lesion localization guided by linguistic cues and interferes with the segmentation of target lesions in the visual environment.
% The efficacy of this design will be validated in the Ablation Study (Sec.~IV.E) and further elucidated through visualization analysis (Sec.~IV.G) to comprehend the disparities and characteristics of features from different stages.
The efficacy of this design will be validated in the Ablation Study (Sec.~IV.E). Further visualization analysis (Sec.~IV.F) will be conducted to explore the disparities and characteristics of features from different stages.

\subsection{Loss Function}

We  design a specialized loss function to constrain the model’s training, which includes the Segmentation Loss $\mathcal{L}_{Seg}$ for the segmentation results and the Vision-Language Contrastive Loss $\mathcal{L}_{Con}$ for the linguistic features and final multi-modal features. 

% \begin{align}
% \mathcal{L}_{P2P} &= - \frac{1}{|P|} \sum_{i}^{|P|} \frac{\text{CS}(p_i, p_{avg})}{\text{CS}(p_i, p_{avg}) + \sum_{j}^{|N|} \text{CS}(p_i, n_j)} \notag \\
% &- \lambda \cdot \frac{1}{|B_P|} \susm_{i}^{|B_P|} \frac{\text{CS}(p_i, p_{avg})}{\text{CS}(p_i, p_{avg}) + \sum_{i}^{|B_N|} \text{CS}(p_i, n_j)} 
% \end{align}
Specifically, we perform dilation and erosion operations on the ground truth label $M^{gt}$ to obtain the boundary region $E$, defined as $E= dilate(M^{gt}) - erode(M^{gt})$. Here, $dilate(\cdot)$ expands and $erode(\cdot)$ contracts the object's boundaries.
A weight $\lambda$ is then assigned to the edge region where $E^i=1$. The loss function for the segmentation mask is defined as:
\begin{align}
\mathcal{L}_{Seg} =  \mathcal{L}_{focal}(M, M^{gt}) + W \odot \mathcal{L}_{dice}(M, M^{gt}),
\end{align}
\noindent where $W^i=\lambda E^i+(1-E^i)$, $\mathcal{L}_{focal}$ and $\mathcal{L}_{dice}$ are focal loss \cite{ross2017focal} and dice loss \cite{milletari2016v}.
% \begin{align}
% \mathcal{L}_{Con} = & - \frac{1}{|P|} \sum_{i}^{|P|} \frac{\mathcal{S}_{cos}(p_i, p_{avg})}{\mathcal{S}_{cos}(p_i, p_{avg}) + \sum_{j}^{|N|} \mathcal{S}_{cos}(p_i, n_j)}  
% % \notag \\
% \end{align}
% \begin{align}
% \mathcal{L}_{Con} &= - \frac{1}{|P|} \sum_{i}^{|P|} \frac{cos(p_i, T_{avg})}{cos(p_i, T_{avg}) + \sum_{j}^{|N|} cos(p_i, n_j)} 
% % &- \beta \cdot \frac{1}{|B_P|} \sum_{i}^{|B_P|} \frac{\text{CS}(p_i, T_{avg})}{\text{CS}(p_i, T_{avg}) + \sum_{j}^{|B_N|} \text{CS}(p_i, n_j)}
% \end{align}

% Meanwhile, $\mathcal{L}_{Con}$ further aligns the visual knowledge of the target region with the linguistic information, leading to improved accuracy in target localization.
Furthermore, $\mathcal{L}_{Con}$ aligns the visual knowledge of the target region with the linguistic information while simultaneously repelling the visual representations of the target and non-target regions, as illustrated in Figure 3(b). In the figure, $S_{align}$ measures the similarity between the features of positive pixels and the average linguistic feature, while $S_{repel}$ quantifies the dissimilarity between the features of positive and negative pixels. The formula for $\mathcal{L}_{Con}$ is as follows:
\begin{align}
\mathcal{L}_{Con} &= - \frac{1}{|\mathcal{P}|} \sum_{i}^{|\mathcal{P}|} \frac{e^{(p_i \cdot L_{avg}/\tau)}}{e^{(p_i \cdot L_{avg} / \tau)} + \sum_{j}^{|\mathcal{N}|} e^{(p_i \cdot n_j/\tau )}},
% &- \beta \cdot \frac{1}{|B_P|} \sum_{i}^{|B_P|} \frac{\text{CS}(p_i, T_{avg})}{\text{CS}(p_i, T_{avg}) + \sum_{j}^{|B_N|} \text{CS}(p_i, n_j)}
\end{align}
% \begin{align}
% \mathcal{L}_{Con} &= - log( \frac{e^{(p_i \cdot T_{avg}/\tau)}}{e^{(p_i \cdot T_{avg} / \tau)} + \sum_{j}^{|N|} e^{(p_i \cdot n_j/\tau )}} )
% % &- \beta \cdot \frac{1}{|B_P|} \sum_{i}^{|B_P|} \frac{\text{CS}(p_i, T_{avg})}{\text{CS}(p_i, T_{avg}) + \sum_{j}^{|B_N|} \text{CS}(p_i, n_j)}
% \end{align}
% \begin{align}
% \mathcal{L}_{Con} &= - \frac{1}{|P|} \sum_{i}^{|P|} \frac{\mathcal{S}_{cos}(p_i, T_{avg})}{\mathcal{S}_{cos}(p_i, T_{avg}) + \sum_{j}^{|N|} \mathcal{S}_{cos}(p_i, n_j)} 
% % &- \beta \cdot \frac{1}{|B_P|} \sum_{i}^{|B_P|} \frac{\text{CS}(p_i, T_{avg})}{\text{CS}(p_i, T_{avg}) + \sum_{j}^{|B_N|} \text{CS}(p_i, n_j)}
% \end{align}
\noindent where $p_i$ represents the feature of the $i$-th positive pixel in the positive pixel set $\mathcal{P}$, $n_j$ denotes the feature of the $j$-th negative pixel in the negative pixel set $\mathcal{N}$, and $\tau$ is a hyperparameter that controls the sharpness of the probability distribution. $L_{avg}$ denotes the average pooled linguistic feature, computed as $L_{avg} = {proj}\left( \frac{1}{T} \sum_{}^{T} L_t \right)$.
$L_t$ is the $t$-th linguistic token.

Our joint  training loss is defined as follows:
\begin{equation}
\mathcal{L} = \alpha\mathcal{L}_{Con} + \mathcal{L}_{Seg},
% + \lambda\mathcal{L}_{Bd.}
\end{equation}
\noindent where $\alpha$ is a  hyperparameter.

\begin{table*}[t]
% \tiny
\centering
\caption{Experimental results on the RefHL-Seg, QaTa-COV19 and MosMedData+ datasets in terms of Dice and mIoU. 
The best scores are in \textcolor{red}{red}, and the secondbest scores are in \textcolor{blue}{blue}.
}
% \label{label1}
\renewcommand{\floatpagefraction}{.9}

\resizebox{\textwidth}{!}{
\renewcommand\arraystretch{1.4}
\begin{tabular}{lcccccccccc}
\toprule
                         &                            &                        &                             &                             & \multicolumn{2}{c}{RefHL-Seg} & \multicolumn{2}{c}{QaTa-COV19} & \multicolumn{2}{c}{MosMedData+} \\ \cline{6-11}
\multirow{-2}{*}{Method} & \multirow{-2}{*}{Backbone} & \multirow{-2}{*}{Text} & \multirow{-2}{*}{Param (M)} & \multirow{-2}{*}{Flops (G)} & Dice (\%)     & mIoU (\%)     & Dice (\%)      & mIoU (\%)     & Dice (\%)      & mIoU (\%)      \\ \hline
 
U-Net  \cite{ronneberger2015u}          & CNN                        & \ding{55}                     & 14.8                        & 50.3                        & 48.67         & 37.89         & 79.02          & 69.46         & 64.60          & 50.73          \\
 
UNet++ \cite{zhou2019unet++}          & CNN                        & \ding{55}                     & 74.5                        & 94.6                        & 51.84         & 40.27         & 79.62          & 70.25         & 71.75          & 58.39          \\
 
AttUNet \cite{oktay2018attention}         & CNN                        & \ding{55}                     & 34.9                        & 101.9                       & 50.67         & 40.11         & 79.31          & 70.04         & 66.34          & 52.82          \\
 
nnUNet \cite{isensee2021nnu}          & CNN                        & \ding{55}                     & 19.1                        & 412.7                       & 51.63         & 41.89         & 80.42          & 70.81         & \textcolor{blue}{72.59}          & 60.36          \\
 
TransUNet \cite{chen2021transunet}       & Hybrid                     & \ding{55}                     & 105                         & 56.7                        & 52.33         & 44.76         & 78.63          & 69.13         & 71.24         & 58.44          \\
 
Swin-Unet \cite{cao2022swin}       & Hybrid                     & \ding{55}                     & 82.3                        & 67.3                        & 51.34         & 44.26         & 78.07          & 68.34         & 63.29          & 50.19          \\
 
UCTransNet \cite{wang2022uctransnet}      & Hybrid                     & \ding{55}                     & 65.6                        & 63.2                        & \textcolor{blue}{54.87}         & 44.38         & 79.15          & 69.60         & 65.90         & 52.69          \\ 
 
LViT \cite{li2023lvit}                   & Hybrid                     & \ding{55}                     & 28.0                        & 54.0                        & 54.84         & \textcolor{blue}{44.65}         & \textcolor{blue}{81.12}          & \textcolor{blue}{71.37}         & {72.58}          & \textcolor{blue}{60.40}          \\ \hline
 
% LSMS (w/o FSD)            & Hybrid                     & \times                     & 28.0                        & 54.0                        & 80.35         & 70.74         & 80.35          & 70.74         & 80.35          & 70.74          \\
 
LSMS                   & Hybrid                     & \ding{55}                     & 8.78                        & 8.91                        & \textcolor{red}{58.75}    & \textcolor{red}{49.39}    & \textcolor{red}{82.14}         & \textcolor{red}{72.07}                & \textcolor{red}{73.14}          & \textcolor{red}{61.76}          \\
\midrule
 
ConVIRT \cite{zhang2022contrastive}        & CNN                        & \ding{51}                      & 35.2                        & 44.6                        & 61.37         & 53.56         & 79.72          & 70.58         & 72.06 & 59.73         \\
 
TGANet \cite{tomar2022tganet}         & CNN                        & \ding{51}                      & 19.8                        & 41.9                        & 63.28         & 55.49         & 79.87          & 70.75         & 71.81 & 59.28         \\
 
GLoRIA \cite{huang2021gloria}          & Hybrid                     & \ding{51}                      & 45.6                        & 60.8                        & 63.69         & 55.82         & 79.94          & 70.68         & 72.42   & 60.18         \\
 
ViLT \cite{kim2021vilt}            & Hybrid                     & \ding{51}                      & 87.4                        & 55.9                        & 64.58         & 56.65         & 79.63          & 70.12         & 72.36   & 60.15         \\

CLIP \cite{radford2021learning}          & Hybrid                     & \ding{51}                      & 87.0                        & 105.3                       & 68.76         & 60.33         & 79.81          & 70.66         & 71.97 & 59.64     \\
 
LAVT \cite{yang2022lavt}            & Hybrid                     & \ding{51}                      & 118.6                       & 83.8                        & 69.03         & 60.98         & 79.28          & 69.89         & 73.29 & 60.41        \\
SLViT \cite{ouyang2023slvit}            & Hybrid                     & \ding{51}                      & 103.4                       & 50.1                        & 71.94         & 62.83         & 80.68          & 71.57         & 73.08 & 60.11        \\
Prompt-RIS \cite{shang2024prompt}   & Hybrid                     & \ding{51}                      & 225.9                       & 108.7                        & 72.11         & 63.08         & 81.13          & 71.94         & 74.12 & 60.97        \\
LViT \cite{li2023lvit}             & Hybrid                     & \ding{51}                      & 29.7                        & 54.1                        & 71.48         & 62.02         & {83.66} & {75.11}        & 74.57 & 61.33        \\ 

RecLMIS \cite{huang2024cross}  & CNN                     & \ding{51}                      & 23.7 & 24.1 & 73.01 & 63.59 &  \textcolor{blue}{85.22} & 77.00 & 77.48 & 65.07              \\ 

 \hline
 
% LSMS (w/o FSD)              & Hybrid                     & \checkmark                      & 8.78                        & 8.91           & 82.73         & 73.99         & 82.73          & 73.99         & 82.73          & 73.99          \\ 
 
LSMS (1/4)                    & Hybrid                     & 25\%                     & 8.78                        & 8.91           &  77.63         &  69.02        &  84.97          &  76.98         &  77.46          &  66.02   \\
LSMS (1/2)                    & Hybrid                     & 50\%                     & 8.78                        & 8.91           &  \textcolor{blue}{78.33}         &  \textcolor{blue}{69.76}        &  {85.21}          &  \textcolor{blue}{77.08}         &  \textcolor{blue}{77.65}          &  \textcolor{blue}{66.24}   \\
LSMS                    & Hybrid                     & 100\%                      & 8.78                        & 8.91          &  \textcolor{red}{78.24}         &  \textcolor{red}{70.31}        &  \textcolor{red}{85.57}          &  \textcolor{red}{77.60}         &  \textcolor{red}{78.32}          &  \textcolor{red}{67.41}   \\

\bottomrule
\end{tabular}
}
\label{tab2}
\vspace{-5pt} 
\end{table*}

\section{Experiments}

\subsection{Datasets}

We conducted experiments on three datasets to assess the performance of LSMS: our self-established dataset, \textbf{\emph{Reference Hepatic Lesion Segmentation (RefHL-Seg)}}, as well as MosMedData+ \cite{morozov2020mosmeddata,hofmanninger2020automatic} and QaTa-COV19 \cite{degerli2022osegnet} datasets. 

RLS, as a new task, lacks suitable benchmarks. 
The benchmarks established for \emph{Natural Image Referring Segmentation} cannot be directly transferred to the medical domain due to 
fundamental differences in acquisition conditions and visual properties.  
Medical images differ from natural images in texture, contrast, and annotation complexity, requiring expert knowledge. Their targets often have unclear boundaries and varying sizes, necessitating multi-scale processing.
% We analyze existing methods and task characteristics, identifying two key challenges in RLS: (i) Robust vision-language modeling, where the unique texture and contrast of medical images necessitate balancing global vision-language relationships with local visual details to ensure precise object localization; (ii) Comprehensive multi-scale interaction, as the varying sizes and unclear boundaries of medical objects make inadequate cross-scale information exchange a potential barrier to accurate lesion boundary identification.
To tackle these issues, we developed the Reference Hepatic Lesion Segmentation (RefHL-Seg) dataset, specifically designed for the RLS task. 
% With radiologists' assistance, we annotated abdominal CT images with liver lesions, providing structured language descriptions of their morphology, location, and type. Corresponding segmentation masks were  delineated. 

% RefHL-Seg stands as the inaugural dataset tailored explicitly for the RLS task, encompassing 2,386 abdominal CT slices from 231 cases. 
As the first dataset tailored specifically for  RLS, RefHL-Seg comprises 2,283 abdominal CT slices from 231 cases, predominantly featuring lesions such as Hemangiomas, Liver Cysts, Hepatocellular Carcinomas (HCC), Focal Nodular Hyperplasia, and Metastasis.
With collaboration from radiology experts, we meticulously annotated all liver lesions for the first time, providing detailed descriptions of their locations and morphologies. Each lesion's textual annotation includes information such as liver segment (location), diameter, shape, boundary, enhancement pattern, lesion type, and more. 
Corresponding segmentation masks were  delineated. 
% For instance, \emph{"In segment VI of the liver, there is a vascular tumor measuring 7.5mm in diameter, exhibiting a regular shape, clear boundary, and non-ring enhancement."}
An exemplary textual annotation containing all relevant information is provided as follows: \emph{``A vascular tumor in  segment VI of the liver, with a diameter of 7.5mm, irregular shape, clear boundary, and non-ring enhancement."} Additionally, experiments will involve testing language expressions that contain only partial information sufficient for lesion localization, such as \emph{"A vascular tumor in segment VI of the liver, with an irregular shape."}

The MosMedData+ \cite{morozov2020mosmeddata,hofmanninger2020automatic} and QaTa-COV19 \cite{degerli2022osegnet} datasets are established  for Classical Lesion Segmentation tasks. Recent study \cite{li2023lvit} extended the textual annotations of these datasets, 
% providing descriptions of the number and regions of all lesions in each slice, 
serving for the evaluation of Text-Augmented Lesion Segmentation. 
The MosMedData+ dataset comprises 2,729 CT scan slices of lung infections, primarily containing information about the location of lung infections and the number of infected regions.
% The MosMedData+ dataset includes 2,729 CT scan slices of pulmonary infections. 
% The QaTa-COV19 dataset, developed by researchers from Qatar University and Tampere University, comprises 9,258 chest X-ray images manually annotated for COVID-19 lesions.
The QaTa-COV19 dataset consists of 9,258 chest X-ray images manually annotated with COVID-19 lesions, focusing on whether both lungs are infected, the quantity of lesions, and the approximate location of the infected regions.

\subsection{Evaluation Metric}

In line with prior research \cite{li2023lvit}, we utilize the Dice score and mean Intersection-over-Union (mIoU) to assess the effectiveness of our proposed method.
% The Dice score measures the similarity between two sets by computing twice the intersection divided by the sum of the sizes of the two sets. 
The Dice score quantifies performance by calculating the intersection of the predicted results and the ground truth, divided by twice the sum of the sizes of the two sets.
% The mIoU calculates the ratio of the intersection to the union of the predicted and ground truth segmentation masks, providing a measure of segmentation accuracy.
The mIoU computes the average Intersection over Union for multiple segmentation instances, providing an overall measure of segmentation accuracy across all instances.

\begin{table}[t]
\centering
% 使用 \captionof 命令为非浮动体添加标题
\caption{Ablation studies on the RefHL-Seg val set. The optimal scores are highlighted in \textcolor{red}{red}.}
\smallskip

\renewcommand{\floatpagefraction}{.9}
\label{tab:shared}

\begin{minipage}{0.48\textwidth}
\centering
\newcommand{\intermediatesize}{\fontsize{8pt}{9.6pt}\selectfont}
    % \footnotesize
    % \small
    \intermediatesize
    \renewcommand\arraystretch{1.3}
% 在这里添加第一个表格的内容
\begin{tabular*}{\textwidth}{@{\extracolsep{\fill}}cccccc}
    \toprule
                           &                            &                            & \multicolumn{1}{c|}{}                          & Dice                 & mIoU                 \\ 
                           \midrule

\multicolumn{6}{l}{(a) Kernel size of the single ConvUnit in SVLA}                                                                                                             \\ \hline

\multicolumn{4}{c|}{$d_1=5$}                                                                                                           & 72.95 & 64.14                           \\
\multicolumn{4}{c|}{$d_1=7$}                                                                                                     &  \textcolor{red}{73.28} & \textcolor{red}{64.81}                           \\ 
\cmidrule(lr){1-6}
\multicolumn{4}{c|}{$d_2=11$}                                                                                                          &  \textcolor{red}{74.03} &{65.27} \\
\multicolumn{4}{c|}{$d_2=15$}                                                                                                          &  73.97    &    \textcolor{red}{65.31}                             \\ 
\cmidrule(lr){1-6}
\multicolumn{4}{c|}{$d_3=19$}                                                                                                          &   72.33   &   63.97                             \\
\multicolumn{4}{c|}{$d_3=21$}                                                                                                          &  \textcolor{red}{73.36} &  \textcolor{red}{65.22}                          \\ 
\midrule

\multicolumn{6}{l}{(b) Ablation on design choices of SVLA}                                                                                                                 \\ \hline
ConvU1                 & ConvU2                 & ConvU3                 & \multicolumn{1}{c|}{PM}      &      &                                \\
\checkmark & \checkmark &                           & \multicolumn{1}{c|}{}                          & 76.75 & 67.34                           \\
\checkmark &                           & \checkmark & \multicolumn{1}{c|}{}                          & 76.89 & 67.45                           \\
                          & \checkmark & \checkmark & \multicolumn{1}{c|}{}                          & 76.65 & 67.22                           \\
\checkmark & \checkmark & \checkmark & \multicolumn{1}{c|}{}                          &   77.23   &     68.91                           \\
\checkmark & \checkmark & \checkmark & \multicolumn{1}{c|}{\checkmark} & \textcolor{red}{78.34} & \textcolor{red}{70.31} \\ 
\midrule
\multicolumn{6}{l}{(c) Effectiveness of FSD}                                                                                                                  \\ \hline
\multicolumn{4}{c|}{LSMS (w/o FSD)}                                                                                                & 74.92 & 65.77                           \\
% \multicolumn{4}{c|}{$P_1, P_2, P_3, P_4$-Head-Concat-MLP}                                                                           & 81.5 & 91.3                           \\
\multicolumn{4}{c|}{LSMS (w/ FSD)}                                                                           & \textcolor{red}{78.34} & \textcolor{red}{70.31}                          \\ \hline
\multicolumn{6}{l}{(d) Ablation on design choices of FSD}                                                                                                                  \\ \hline
\multicolumn{4}{c|}{$P_4$-Head-MLP}                                                                                                & 76.03 & 67.58                             \\
% \multicolumn{4}{c|}{$P_1, P_2, P_3, P_4$-Head-Concat-MLP}                                                                           & 81.5 & 91.3                           \\
\multicolumn{4}{c|}{$P_1, P_2, P_3, P_4$-MLP-Concat-MLP}                                                                           & 77.41 & 68.79                          \\
\multicolumn{4}{c|}{$P_1, P_2, P_3, P_4$-Concat-Head-MLP}                                                                          & \textcolor{red}{77.97} & \textcolor{red}{69.38} \\ 
\multicolumn{4}{c|}{$P_1, P_2, P_3, P_4$-MLP-Concat-Head-MLP}                                                                           & 76.40 & 68.29                           \\ \midrule
\multicolumn{6}{l}{(e) FSD on various stages}                                                                                                                              \\ \hline
$P_1\ \ \ \  $ & $P_2\ \ \ \ \ \ \ \  $ & $P_3\ $ & \multicolumn{1}{c|}{$P_4$} &      &                                \\
                          &                           &                           & \multicolumn{1}{c|}{\checkmark} & 77.94 & 66.59                           \\
                          &                           & \checkmark & \multicolumn{1}{c|}{\checkmark} & 78.80 & 67.54                           \\
\checkmark & \checkmark & \checkmark & \multicolumn{1}{c|}{}                          & 77.89 & 67.36                           \\
                          & \checkmark & \checkmark & \multicolumn{1}{c|}{\checkmark} & \textcolor{red}{78.34} & \textcolor{red}{70.31}                         \\
\checkmark & \checkmark & \checkmark & \multicolumn{1}{c|}{\checkmark} & 77.97 & 69.38  \\ 
    \bottomrule
    % \end{tabular}
    \end{tabular*}
\end{minipage}
\label{tab8}
% \vspace{-10pt}
\end{table}

% \vspace{-6pt}
\subsection{Implementation Details}

The proposed LSMS is implemented using PyTorch, leveraging the BERT implementation from the HuggingFace Transformer library \cite{wolf2020Transformers}. Regarding dataset partitioning, we separated the original training set into training and validation sets. 
% The sample sizes for each dataset are listed in Tab.~\ref{tab8}.
For the convolutional layers in the Visual Knowledge Branch of SVLA, we initialized the weights using SegNeXt \cite{NEURIPS2022_08050f40} pre-trained on ImageNet-22K \cite{guo2022segnext}. The language encoder of LSMS was initialized with the official pre-trained BERT weights, consisting of 12 layers with a hidden size of 768. 
The number of encoder blocks and feature dimensions for each stage are presented in Tab.~\ref{tab1}. For the convolutional branch of SVLA, we used kernel sizes of $d_1 = 7$, $d_2 = 11$, and $d_3 = 21$. 
We set the default values for key hyperparameters as follows: $\alpha=0.125$, $\lambda=1.2$ and $\tau = 0.1$. 
% Additional hyperparameter settings and experiments are detailed in the Appendix.
The remaining weights in our model were randomly initialized. 

Subsequently, we employed the AdamW optimizer with a weight decay of 0.01. The learning rate was initialized to 3e-5 and scheduled using polynomial decay with a power of 0.9. All models were trained for 100 epochs with a batch size of 16. Images were uniformly resized to $480 \times 480$ before inputting into the model, with no additional data augmentation applied.

\subsection{Comparison with the State-of-the-Arts}

\paragraph{\textbf{Classical Lesion Segmentation}}
We compared the performance of LSMS with existing segmentation models on the RefHL-Seg, QaTa-COV19, and MosMedData+ datasets, as shown in Tab.~\ref{tab2}. 
In scenarios where  Text is not utilized, 
we remove the language-related components from LSMS, such as the Language-Guided Branch in SVLA and the $\mathcal{L}_{Con}$ term in the loss function.
We observed that LSMS outperformed all existing methods in Classical Lesion Segmentation task while minimizing computational costs. 
% LSMS outperforms all existing methods in classical medical image segmentation task.
% On the RefHL-Seg dataset, LSMS attains a Dice score of 58.75\% and an mIoU score of 49.39\%, outperforming all other models. On the QaTa-COV19 dataset, LSMS achieves a Dice score of 82.14\% and an mIoU score of 72.07\%, ranking first in both metrics. On the MosMedData+ dataset, LSMS achieves the highest Dice score of 73.14\% and the best mIoU score of 61.76\%.
This underscores the efficiency and superiority of LSMS in understanding medical visual environment.
% While achieving higher segmentation accuracy, LSMS minimizes computational costs. 
% Its ability to outperform more complex models while maintaining a lightweight architecture underscores its potential for practical applications in clinical and research settings.

\paragraph{\textbf{Text-Augmented Lesion Segmentation}}
The QaTa-COV19 and MosMedData+ datasets have been extended with textual annotations for training and validating Text-Augmented Lesion Segmentation. In Tab.~\ref{tab2}, our LSMS achieves state-of-the-art (SOTA) performance.
As the proportion of textual input increases, LSMS consistently improves across all evaluation metrics.
% , with LSMS (100\% Text) achieving 85.57\% Dice and 77.60\% mIoU on QaTa-COV19, and 78.32\% Dice and 67.41\% mIoU on MosMedData+. 
% Particularly, LSMS  improving Dice by 3.21\% and mIoU by 5.50\% on the MosMedData+ dataset. 
% These results demonstrate that incorporating textual descriptions helps LSMS to extract more informative features, leading to enhanced segmentation accuracy compared to models relying solely on visual input.
This indicates that LSMS, with textual assistance, can complement visual information with linguistic information, leading to a more accurate segmentation of the lesions.

\paragraph{\textbf{Referring Lesion Segmentation}}
The RefHL-Seg dataset we constructed serves the specific purpose of training and validating the RLS task. Each sample in the dataset necessitates the model to segment the particular lesion indicated by the reference expression, thus rendering the performance of all methods suboptimal when text input is absent. 
% When provided with complete input comprising both text and image, our LSMS significantly improves accuracy compared to all existing vision-language segmentation models. This improvement stems from the inadequacy of existing methods to adapt to the complexity of medical visual environment or accurately locate specific lesions based on linguistic cues. 
% Specifically, compared to the existing vision-language models LViT \cite{li2023lvit} and RecLMIS \cite{huang2024cross}, LSMS achieves an increase of 8.29\%  and 6.72\% in mIoU.
Our LSMS serves as a novel baseline for the RLS task, demonstrating its significant advantages over methods applied to previous related tasks. To provide a comprehensive comparison, we evaluate LSMS against existing vision-language models on the newly proposed task and dataset. 
When provided with complete inputs containing both text and images, our LSMS significantly improves accuracy compared to all existing models.
% The results show that LSMS outperforms them by achieving a substantial improvement. 
Specifically, LSMS achieves an increase of 8.29\% over LViT \cite{li2023lvit} and 6.72\% over RecLMIS \cite{huang2024cross} in mIoU, highlighting its effectiveness and superior performance in this new benchmark.
% Specifically, compared to existing vision-language models, LViT \cite{li2023lvit} and RecLMIS \cite{huang2024cross}, the proposed LSMS demonstrates a significant improvement in mIoU, with increases of 8.29\% and 6.72\%, respectively.
This improvement stems from LSMS's enhanced understanding of the complex medical visual environment and its precise localization capability based on linguistic cues.
% In Tab.~\ref{tab2}, LSMS (1/4) and LSMS (1/2) respectively denote the utilization of only 25\% and 50\% of the text for testing, indicating that LSMS's performance still notably surpasses existing methods. 
In Tab.~\ref{tab2}, LSMS (1/4) and LSMS (1/2) denote the models tested with only 25\% and 50\% of the textual input, respectively, by omitting portions such as size or shape descriptions. The results indicate that LSMS still outperforms existing methods significantly.
This underscores that our LSMS, trained to precisely locate specified lesions in images with minimal textual guidance, exhibits outstanding language-guided lesion localization capability.
\\
\\

\begin{figure*}[ht!]
    \centering
    \includegraphics[width=1.0\linewidth]{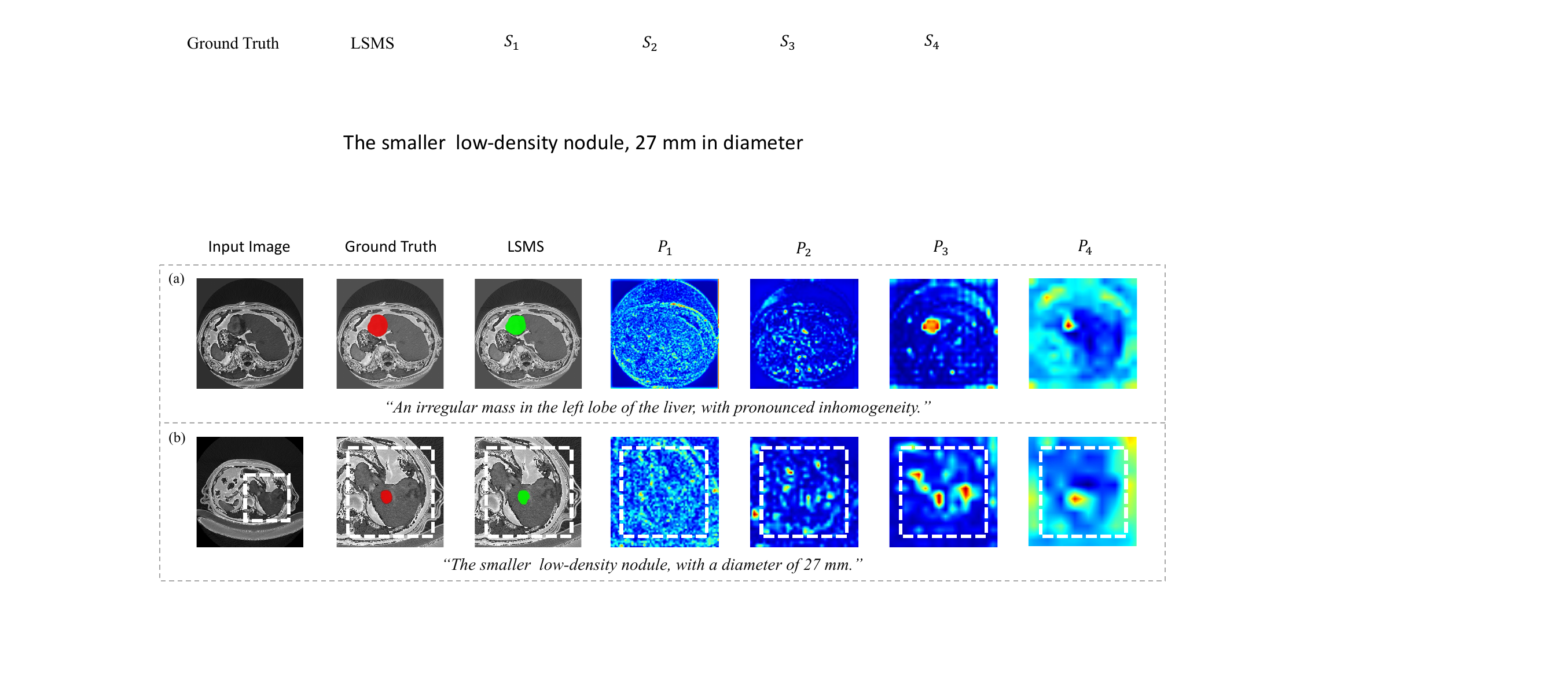}
    \vspace{-15pt}
    \caption{Visualization of the feature maps from different stages in LSMS.
    The red regions denote the Ground Truth, while the green regions represent the segmentation results of our LSMS.
    In sample (b), for ease of observation, the key region within the image have been enlarged, with a white rectangular box serving as a reference for location.
    }
    \label{fig4}
    \vspace{-10pt}
\end{figure*}

\vspace{-25pt}
\subsection{Ablation Study}

\paragraph{\textbf{Kernel size of the single ConvUnit in SVLA}}
We have meticulously evaluated the performance of using different convolutional kernels in SVLA on the validation set of RefHL-Seg dataset, as illustrated in Tab.~\ref{tab8}(a). $d_j=N$ indicates the ConvUnit containing a $1 \times N$ convolution and a $N \times 1$ convolution. The strip-like convolution kernels aim to obtain detailed local visual information with low costs. 
We employed a single ConvUnit in SVLA to evaluate the impact of various convolution kernel sizes $d_j$. In Tab.~\ref{tab8}(a), sizes 7 and 21 show superior performance among those with comparable computational costs.
The performance of medium-size convolution kernels is similar, so we choose size 11 due to its lower computational cost.
The utilization of diverse convolution kernels can aid in capturing features from varying receptive fields, 
which helps in extracting rich local visual features.
Therefore, we selected  kernel sizes of $d_1=7$, $d_2=11$, and $d_3=21$ as the default settings.

\paragraph{\textbf{Ablation on SVLA design}}
In Tab.~\ref{tab8}(b), ConvUnit$j$ comprises a $1 \times d_j$ convolution and a $d_j \times 1$ convolution, Pixel-Map (PM) represents a element-wise matrix multiplication operation in the Language-Guided Branch.
% aimed at enhancing language-guided visual knowledge.
Tab.~\ref{tab8}(b) shows that the deployment of three ConvUnits produces the most favorable outcomes, and
the incorporation of the Pixel-Map enhances language-guided visual knowledge, significantly boosting segmentation accuracy.

\paragraph{\textbf{Effectiveness of FSD}}
To assess the effectiveness of the FSD, 
% we conducted ablation studies on three benchmark datasets (i.e., RefHL-Seg, XX, and XX). Specifically, 
we compared the performance of the complete LSMS with LSMS (w/o FSD), and the results are reported in Tab.~\ref{tab8}(c). Experimental findings indicate that removing the FSD module leads to performance degradation, with Dice decrease of 3.42\% and mIoU decrease of 4.54\%, respectively. 
% Furthermore, examining the parameter and GFLOPs presented in Tab.~\ref{tab2} reveals that despite a slight increase in parameters and GFLOPs, FSD enhances performance, underscoring its efficiency in improving performance through global modeling of multi-modal information.
Furthermore, FSD incurs only a modest increase of 2.1M parameters and 1.7G Flops, yet significantly enhances performance. This underscores its efficiency in boosting performance by globally modeling multi-modal information.

\paragraph{\textbf{Ablation on FSD design}}
To validate the effectiveness of the individual components in the design of FSD, we conducted ablation experiments on its constituents, as presented in Tab.~\ref{tab8}(d). It is evident that relying solely on features from the final stage is insufficient, and while adding MLP layers before concatenation increases complexity, it results in the loss of valuable information from each stage. The incorporation of the Hamburger Head design enhances the ability to globally model multi-scale information, consequently improving segmentation performance. The $P_1, P_2, P_3, P_4$-Concat-Head-MLP design demonstrates the best performance.

\paragraph{\textbf{FSD on various stages}}
Given the multi-modal features from different stages after Position Alignment, FSD concatenates them and performs joint refinement in a single forward pass. Here, $P_i$,  $i \in \{1,2,3,4\}$, denotes the multi-modal features input to FSD from the $i$-th stage. Tab.~\ref{tab8}(e) compares multiple input sequences, confirming the value of multi-scale interaction for global reasoning. As shown, $P_2, P_3, P_4$ exhibit optimal performance for FSD.
This superiority is attributed to the insufficient depth of interaction between visual information and linguistic cues in the shallow feature $P_1$, which include irrelevant information  and hindering the localization and segmentation of specified lesions. Conversely, the latter three layers demonstrate strong visual-linguistic consistency, facilitating favorable predictions.
Visualization analysis of each feature map is detailed in Sec.~IV.F.

% Tab.~\ref{tab4}  summarizes the results of the experiments, where we tested the following values for $\alpha$: 0.08, 0.1, 0.2, 0.11, 0.12, 0.125, and 0.13. The Dice scores and mIoU values corresponding to each $\alpha$ value are shown, with the optimal scores highlighted in red.

% The results indicate that the best performance was achieved with $\alpha$ = 0.125, yielding a Dice score of 78.02 and an mIoU of 69.66. These values represent the highest segmentation accuracy and overlap among the tested configurations. In comparison, the other values of $\alpha$, such as 0.08 and 0.2, resulted in lower scores for both metrics, with Dice scores ranging from 76.07 to 77.85 and mIoU values between 67.52 and 69.44.

% Based on these findings, $\alpha$ = 0.125 is chosen as the optimal value, as it provides the most favorable performance in terms of both Dice and mIoU. This value will be used in subsequent experiments and analyses.

% \vspace{-6pt}
\subsection{Visualization Analysis}

In Fig.~\ref{fig4}, we illustrate segmentation results and feature maps obtained from two pairs of inputs, denoted as (a) and (b). In Fig.~\ref{fig4}(a), the language expression is “An irregular mass in the left lobe of the liver, with pronounced inhomogeneity.”   The language expression for Fig.~\ref{fig4}(b) is “The smaller  low-density nodule, 27 mm in diameter.” The labels $P_i, i\in \{1,2,3,4\}$ represent encoded multi-modal features from different stages. 
The segmentation results demonstrate LSMS's ability to accurately locate and segment the specified lesions based on the language expressions. Analyzing $P_i, i\in \{1,2,3,4\}$ reveals LSMS's progressive focus from shallow to deep layers onto the corresponding lesions: $P_1$ comprehends the overall visual context, $P_2$ extensively attends to various objects within the image, $P_3$ narrows down to candidate lesions, and $P_4$ precisely focuses on the lesion specified by the language input.
In sample (a), where only one large lesion is present, $P_3$ rapidly focuses on the target area. As modeling deepens, $P_4$ demonstrates profound visual-linguistic cues. In sample (b), where multiple lesions coexist in the image, $P_3$ exhibits multiple attention points.
% , and through further visual-language cross-modal interaction, $P_4$ can focus on the language-guided lesions, aiding the model in making correct predictions.
Through further vision-language interaction, $P_4$ can focus on the lesions specified by the expression, aiding the model in making correct predictions.

\section{Conclusion}

In this paper, we introduce a new task, Referring Lesion Segmentation, driven by clinical demands. 
% Additionally, we develop the RefHL-Seg dataset for training and validating the RLS task.
To support this task, we develop the RefHL-Seg benchmark for training and validating  RLS.
% we develop the RefHL-Seg benchmark.
We propose LSMS for lesion segmentation with two appealing designs.
LSMS integrates a novel attention mechanism to enhance object localization by tightly interacting scale-aware visual knowledge and linguistic cues.
We introduce a full-scale decoder for global modeling of multi-modal features, improving boundary prediction in segmentation.
Additionally, we design a specialized loss function to enhance fine-grained discrimination. 
Our experimental results show LSMS outperforms existing methods in both RLS and conventional lesion segmentation tasks with lower computational costs.

\bibliographystyle{unsrt} % 根据需求选择不同的样式
\bibliography{bare_jrnl_new_sample4} % 指定你的BibTeX文件

\vfill

\end{document}